\documentclass{article} 

\usepackage{PRIMEarxiv}

\usepackage[english]{babel}
\usepackage{amsfonts}
\usepackage{amsmath}
\usepackage{amsthm}

\usepackage[utf8]{inputenc}
\usepackage{graphicx}
\usepackage{todonotes}
\usepackage[ruled, linesnumbered, noend]{algorithm2e}
\usepackage{placeins}
\usepackage{subcaption}
\usepackage{rotating}
\usepackage{hyperref}
\usepackage{fancyhdr}
\usepackage{placeins}
\usepackage{float}
\usepackage{booktabs}
\usepackage{multirow}
\usepackage{orcidlink}

\title{Bi-Level Multi-View fuzzy Clustering with Exponential Distance}

\author{Kristina P. Sinaga \thanks{\texttt{kristinapestaria.sinaga@isti.cnr.it}. Istituto di Scienza e Tecnologie dell'Informazione, Italian National Research Council, Italy.}}

\begin{document}
\maketitle
\begin{abstract}
In this study, we propose extension of fuzzy c-means (FCM) clustering in multi-view environments.  First, we introduce an exponential multi-view FCM (E-MVFCM). E-MVFCM is a centralized MVC with consideration to heat-kernel coefficients (H-KC) and weight factors. Secondly, we propose an exponential bi-level multi-view fuzzy c-means clustering (EB-MVFCM). Different to E-MVFCM, EB-MVFCM does automatic computation of feature and weight factors simultaneously. Like E-MVFCM, EB-MVFCM present explicit forms of the H-KC to simplify the generation of the heat-kernel $\mathcal{K}(t)$ in powers of the proper time $t$ during the clustering process. All the features used in this study, including tools and functions of proposed algorithms will be made available at https://www.github.com/KristinaP09/EB-MVFCM.
\end{abstract}


\section{Introduction}

Heat-kernel coefficients (H-KC) are a fundamental tool in quantum field theory (QFT), especially in the context of quantum field theory in curved space-time and the study of quantum anomalies~\cite{vassilevich2003heat}. As renormalization is crucial in machine learning (ML), elaborating H-KC in the systematic process not only identify and quantify the divergences that arise in quantum field theory calculations. But also allowing the computations process to be managed and eliminated to produce finite and meaningful results~\cite{bel1996calculation}. 

Special attention is given to the use of kernels in unsupervised learning, particularly clustering, where they facilitate complexity reduction and efficient data handling through methods like principle component analysis (PCA) and support vector machines (SVM)~\cite{schaback2006kernel}. Kernel PCA is a technique that uses kernels to perform PCA in the transformed feature space. This helps in reducing the dimensionality of the data while preserving the structure necessary for clustering. Kernels behind the principle of SVM is used to define hyperplanes that separate clusters in the feature space. Overall, kernels enhance pattern recognition by providing powerful tools for data transformation, similarity measurement, dimensionality reduction, and regularization, leading to more accurate, efficient, robust recognition of patterns in complex data.

In unsupervised pattern recognition tasks, kernel are used in clustering algorithms like spectral clustering. These algorithms leverage the kernel matrix to identify clusters of similar patterns, even in complex data. Spectral clustering works simply by employing the eigenvalues and eigenvectors of graph Laplacian matrices to find a partition of the graph. The clustering algorithm such as k-means will be applied ones the graph representation, graph Laplacian, eigenvectors computation, and transformation of the original data points into a new representation in a lower-dimensional space accomplished~\cite{von2007tutorial}. This approach mainly leverages the properties of the graph Laplacian to enhance the cluster properties in the data, making it easier to detect clusters in the new representation. This means that the spectral clustering employs several important implications to capture the cluster structure of the data space. In other words, the quality of the clusters obtained from spectral clustering can vary depending on the dataset and the chosen similarity graph.

In unsupervised machine learning, soft clustering is the most powerful method for discovering patterns and various asymptotics of the effective action in solving real-world problems. Here, we will focus on the fuzzy c-means (FCM) method and an explicit expression of heat-kernel coefficients to transform the discretized Euclidean distance between data points and cluster centers. First, we will delve into the mathematical aspects of standard FCM and strictly centered the heat kernel expansion in addressing multi-view problems. Specifically, we will leverage the mathematical structure of H-KC using standard FCM to create new algorithms that can perform inference and learning tasks in multi-view environments.

The remainder of this paper is structured as follows. Section~\ref{sec:background_and_setting} reviews related work, summarizing key findings from existing studies on the theoritical foundations of kernels, the exponential (Gaussian) kernel, and kernel based clustering. Section~\ref{sec:methodolgy} details the proposed objective functions, optimization, algorithms, and the signifiance of the parameter in E-MVFCM and EB-MVFCM.

\section{Background and Setting}
\label{sec:background_and_setting} 
\textbf{The Theoretical Foundations of Kernels:} Kernels are highlighted as essential tools due to their flexibility, optimal recovery capabilities, and ability to handle large-scale data analysis problem. The flexibility and generality in kernels can be tailored to meet the demands of various applications, allowing them to handle diverse types of data inputs, such as texts, images, and more complex structures including multi-view data in federated and non-federated environments. This flexibility makes them suitable for a wide range of data analysis tasks. In their simplest form, kernel-based methods can make optimal use of the given information, providing the best possible approximation or interpolation of functions from unstructured data. This is crucial for accurate data analysis and modeling~\cite{von2007tutorial}. 

In machine learning and pattern recognition, kernels enable the application of linear techniques in high-dimensional feature spaces, even when the original data set does not have a predefined structure. To avoid overfitting and ensure stability in the models, the used of kernels can generate smooth functions and provide regularization. This adaptivity enhances the accuracy and efficiency of data analysis. More importantly, kernel-based methods can avoid numerical integration, which simplifies the computation and reduces potential sources of error~\cite{schaback2006kernel}.

\textbf{The Exponential (Gaussian) Kernel~\cite{schaback2006kernel}:} The exponential kernel, also known as the absolute exponential kernel, has the form:

\begin{equation}
\mathcal{K}_{\exp}(x,y) = \exp \bigg(- \big\|x-y\big\|_2^2 \bigg)  \text{  for all  } x, y \in \Omega := \mathbb{R}^d
\end{equation}

where $\Omega$ serves as the domain or the set of possible inputs for the kernel function. Specifically, $\Omega$ can be an arbitrary nonempty set, and it defines the space in which the kernel operates. In other words, the set $\Omega$ defines the possible learning inputs. The kernel function $\mathcal{K}$ is defined as:
\begin{equation}
\mathcal{K} : \Omega \times \Omega \rightarrow \mathbb{R}
\label{eqn:kernelfunc}
\end{equation}

Eq. \ref{eqn:kernelfunc} means that the kernel takes two elements from $\Omega$ and maps them to a real number. The purpose of $\Omega$ is to provide the context or the space within which the kernel function is applied. In different applications, $\Omega$ can represent various types of data. For instance, in machine learning, $\Omega$ could be the set of all possible feature vectors in machine learning. In image processing, $\Omega$ could be the set of all possible pixel values or image patches. In text analysis, $\Omega$ could be the set of all possible words or documents. In numerical analysis, $\Omega$ could be the set of all possible points in a geometric domain. 

\textbf{Kernel Based Clustering:} Kernels relate to unsupervised clustering in several key ways including feature space transformation, similarity measures, principle component analysis (PCA), support vector machines (SVM), spectral clustering, complexity reduction, and handling non-linear relationships. 

The objective function of spectral clustering aims to partition a graph such that the edges between different groups have very low weights (indicating dissimilarity between points in different clusters) and the edges within a group have high weights (indicating similarity between points within the same cluster). In spectral clustering, the fully connected graph can be approximated to handle large datasets efficiently. The fully connected graph with the Gaussian similarity function is a powerful tool in spectral clustering for capturing local neighborhood relationships. 

In spectral clustering, the Gaussian similarity function is defined as~\cite{von2007tutorial}:

\begin{equation}
s(x_i, x_j) = \exp \bigg( \frac{|x_i - x_j|^2} {2\sigma^2} \bigg)
\label{eqn:GSF}
\end{equation}

where $|x_i - x_j|$ is the Euclidean distance between data points $x_i$ and $x_j$, and $\sigma$ is a parameter controlling the width of the neighborhoods. The Gaussian similarity function in Eq. \ref{eqn:GSF} ensures that points in local neighborhoods are connected with relatively high weights, while edges between far-away points have positive but negligible weights. This helps in capturing the local structure of the data effectively.

The clustering objective of the fuzzy c-means (FCM) procedure is given as follows~\cite{bezdek1984fcm}:

\begin{equation}
J_{FCM}(U,V) = \sum_{i=1}^n \sum_{k=1}^c \mu_{ik}^m \|x_i - a_k\|^2
\label{eqn:fcm}
\end{equation}

As visualize in Eq. \ref{eqn:fcm}, the standard FCM formulating the distance between data points $x_{ij}$ and cluster centers $a_{kj}$ in Euclidean space. The general Euclidean distance is beneficial for soft clustering procedures when input data are continuous and exhibit smooth behavior. In other words, the Euclidean distance $\|x_i - a_k\|^2$ in Eq. \ref{eqn:fcm} is suitable for capturing smooth phenomena.

\section{Methodology}
\label{sec:methodolgy} 

\subsection{Notation}
We assume there are $s$ number of views representing input data $X$, such as a dataset $X \in \mathbb{R}^{n \times D}$ with $X^h = \{ x_1^{1} ,\ldots, x_n^{s} \}_{h=1}^{s}$, $x_i^h = \{x_{1d_1}^{1}, \ldots , x_{nd_s}^s\}_{j=1}^{d_h}$, and $\sum_{h=1}^s d_h = D$. $x_i^{h}$ represents the $i$th points of $x$ in $h$th view. A partition $\mu$ in $X$ is a set of disjoint clusters that partitions $X$ into $c$ clusters across views $h$ such as $\mu^* = [\mu_{ik}^*]_{n \times c}$, $\mu_{ik}^* \in [0, 1]$ and $\sum_{k=1}^c \mu_{ik}^* = 1$. $v$ is a view weight vector with  $v_h \in [0,1]$ and $\sum_{h=1}^s v_h = 1$. We use cluster centers $k$ of $h$th view as $A^h$, where $A^h = [a_{kj}^h]_{c \times d_h}$.

\subsection{The E-MVFCM}

To enable the standard FCM can learn optimal coupling constants that approximates the distribution of feature weights and weight factors in the $h$-th view, we introduce the heat-kernel coefficients (H-KC) for efficient inference in clustering process. In general, the heat-kernel theory $\mathcal{K}$ can progressively extract features of increased abstraction in data to enhance the expressivity and representational capacity. For implementing this architecture in multi-view clustering (MVC), we will reweighting the discritized Eclidean distance $\|x_i - a_k\|^2$ by recasting the asymmetric distance between the data points $x_{ij}$ and cluster centers $a_{kj}$ as exponential kernel distance in a Euclidean space. 

\begin{equation}
J_{\text{E-MVFCM}} (V, U, A) = \sum_{h=1}^s v_h^{\alpha} \sum_{i=1}^n \sum_{k=1}^c (\mu_{ik}^*)^m d_{\text{exp}(ik,j)}^h \label{eqn:E-MVFMC}
\end{equation}

\begin{equation*}
\text{s.t.} \sum_{k=1}^c \mu_{ik}^* = 1, \mu_{ik}^* \in [0,1] \text{~~and~~} \sum_{h=1}^s v_h = 1, v_h \in [0,1]
\end{equation*}

where $d_{\text{exp}(ik,j)}^h = \{1 - \text{exp}(-\sum_{j=1}^{d_h} \delta_{ij}^h (x_{ij}^h - a_{kj}^h)^2 \}$, the kernel coefficient $\delta_{ij}^h = \|x_{ij}^h - \bar{x}^h\|$, and $\bar{x}^h = \frac{x_{ij}^h}{n}$. As visible in Eq. \ref{eqn:E-MVFMC}, the E-MVFCM is directly computed the membership matrix $U$ in a way that it is consistent and shared across all views rather than computing separate memberhips for each view and then combining them. In other words, our proposed E-MVFCM enforces the membership matrices $U$ are the same across different views.

\subsubsection{Optimization}

The optimization of our proposed E-MVFCM involves iteratively updating the "common" memberships $U$, the weight factors $V$, and the cluster centers $A$ until convergence. The Lagrangian of proposed E-MVFCM in Eq.  \ref{eqn:E-MVFMC}  can be expressed as:

\begin{equation}
\tilde{J}_{\text{E-MVFCM}} (V, U, A, \lambda_1, \lambda_2) =J_{\text{E-MVFCM}} (V, U, A)  + \lambda_1 (\sum_{k=1}^c \mu_{ik}^* -1) + \lambda_2 (\sum_{h=1}^s v_h -1)
\label{eqn:LagE-MVFMC}
\end{equation}

\paragraph{Step 1:} The updating function for \textit{"common"} memberships $\mu_{ik}^*$

\begin{align*}
\frac{\partial \tilde{J}_{\text{E-MVFCM}} (V, U, A, \lambda_1, \lambda_2) }{\partial \mu_{ik}^*} & =  m(\mu_{ik}^*)^{m-1} \sum_{h=1}^s v_h^\alpha d_{\text{exp}(ik,j)}^h  + \lambda_1 = 0\\
m(\mu_{ik}^*)^{m-1} \sum_{h=1}^s v_h^\alpha d_{\text{exp}(ik,j)}^h &=  -\lambda_1 \\
\mu_{ik}^*&= \big(\frac{-\lambda_1}{m}\big)^{\frac{1}{(m-1)}} \bigg( \frac{1}{  \sum_{h=1}^s v_h^\alpha d_{\text{exp}(ik,j)}^h} \bigg)^{\frac{1}{(m-1)}}
\end{align*}

Since $\sum_{k=1}^c \mu_{ik}^* = 1$, then we have $\sum_{k'=1}^c \big(\frac{-\lambda_1}{m}\big)^{\frac{1}{(m-1)}} \bigg( \frac{1}{  \sum_{h=1}^s v_h^\alpha d_{\text{exp}(ik',j)}^h} \bigg)^{\frac{1}{(m-1)}} = 1$ and $ \big(\frac{-\lambda_1}{m}\big)^{\frac{1}{(m-1)}} = 1/ \sum_{k'=1}^c \bigg( \frac{1}{\sum_{h=1}^s v_h^\alpha d_{\text{exp}(ik',j)}^h} \bigg)^{\frac{1}{(m-1)}} $. Thus we can update the cluster memberships by using the following equation.

\begin{equation}
\mu_{ik}^* = \frac{ \bigg( \frac{1}{\sum_{h=1}^s v_h^\alpha d_{\text{exp}(ik,j)}^h} \bigg)^{\frac{1}{(m-1)}}} {\sum_{k'=1}^c \bigg( \frac{1}{\sum_{h=1}^s v_h^\alpha d_{\text{exp}(ik',j)}^h} \bigg)^{\frac{1}{(m-1)}}}
\label{eqn:UpdateU_E-MVFMC}
\end{equation}

\paragraph{Step 2:} The updating function for weight factors  $v_h$
\begin{align*}
\frac{\partial \tilde{J}_{\text{E-MVFCM}} (V, U, A, \lambda_1, \lambda_2) }{\partial v_h} & =  \alpha v_h^{\alpha - 1} \sum_{i=1}^n \sum_{k=1}^c (\mu_{ik}^*)^m d_{\text{exp}(ik,j)}^h + \lambda_2 = 0\\
 v_h^{\alpha - 1} \sum_{i=1}^n \sum_{k=1}^c (\mu_{ik}^*)^md_{\text{exp}(ik,j)}^h &=  -\lambda_2 \\
v_h&= \big(\frac{-\lambda_2}{\alpha}\big)^{\frac{1}{(\alpha - 1)}} \bigg( \frac{1}{   \sum_{i=1}^n \sum_{k=1}^c (\mu_{ik}^*)^md_{\text{exp}(ik,j)}^h } \bigg)^{\frac{1}{(\alpha - 1)}}
\end{align*}

Since $\sum_{h=1}^s v_h = 1$, then we have $\sum_{h'=1}^s \big(\frac{-\lambda_2}{\alpha}\big)^{\frac{1}{(\alpha - 1)}} \bigg( \frac{1}{   \sum_{i=1}^n \sum_{k=1}^c (\mu_{ik}^*)^md_{\text{exp}(ik,j)}^{h'} } \bigg)^{\frac{1}{(\alpha - 1)}} =1$ and $\big(\frac{-\lambda_2}{\alpha}\big)^{\frac{1}{(\alpha - 1)}} = \frac{1} {\sum_{h'=1}^s \bigg( \frac{1}{   \sum_{i=1}^n \sum_{k=1}^c (\mu_{ik}^*)^md_{\text{exp}(ik,j)}^{h'} } \bigg)^{\frac{1}{(\alpha - 1)}} }$. Thus, we can obtain an iterative procedure for updating \textit{h}-th view cluster centers as follows.

\begin{equation}
v_h = \frac{\bigg( \frac{1}{   \sum_{i=1}^n \sum_{k=1}^c (\mu_{ik}^*)^md_{\text{exp}(ik,j)}^h } \bigg)^{\frac{1}{(\alpha - 1)}}} {\sum_{h'=1}^s \bigg( \frac{1}{   \sum_{i=1}^n \sum_{k=1}^c (\mu_{ik}^*)^md_{\text{exp}(ik,j)}^{h'} } \bigg)^{\frac{1}{(\alpha - 1)}}}
\label{eqn:UpdateV_E-MVFCM}
\end{equation}

\paragraph{Step 3:} The updating function for the $h$-th view cluster centers $a_{kj}^h$

\begin{align*}
\frac{\partial J_{\text{E-MVFCM}} (V, U, A) }{\partial a_{kj}^h} & = -2 (x_{ij}^h - a_{kj}^h) v_h^\alpha \sum_{i=1}^n  (\mu_{ik}^*)^m \{ - \text{exp}(-\delta_{i}^h \|x_i^h - a_k^h\|^2 \} = 0\\
 v_h^\alpha \sum_{i=1}^n  (\mu_{ik}^*)^m \{ - \text{exp}(-\delta_{i}^h \|x_i^h - a_k^h\|^2 \}  a_{kj}^h&=  v_h^\alpha \sum_{i=1}^n  (\mu_{ik}^*)^m \{ - \text{exp}(-\delta_{i}^h \|x_i^h - a_k^h\|^2 \}  x_{ij}^h \\
 a_{kj}^h&=  \frac{v_h^\alpha \sum_{i=1}^n  (\mu_{ik}^*)^m \{ - \text{exp}(-\delta_{i}^h \|x_i^h - a_k^h\|^2 \} } {v_h^\alpha \sum_{i=1}^n  (\mu_{ik}^*)^m \{ - \text{exp}(-\delta_{i}^h \|x_i^h - a_k^h\|^2 \} }  x_{ij}^h 
\end{align*}

Thus, we can update the cluster centers using the following equation.
\begin{equation}
 a_{kj}^h =  \frac{ \sum_{i=1}^n  v_h^\alpha (\mu_{ik}^*)^m \{ - \text{exp}(-\delta_{i}^h \|x_i^h - a_k^h\|^2 \} } {\sum_{i=1}^n  v_h^\alpha (\mu_{ik}^*)^m \{ - \text{exp}(-\delta_{i}^h \|x_i^h - a_k^h\|^2 \} }  x_{ij}^h 
 \label{eqn:UpdateA_E-MVFCM}
\end{equation}

The algorithm for proposed \texttt{E-MVFCM} in Eq. \ref{eqn:E-MVFMC} is provided as below.

\begin{algorithm}[h]
\caption{The E-MVFCM}
\label{alg:E-MVFCM}
\SetKwInOut{Input}{Input}\SetKwInOut{Output}{Output}
\SetKwProg{ForAllInParallel}{foreach}{ do in parallel}{end} 
\Input{
Multi-view data set $X = \{x_1, x_2, \ldots, x_n\}$ with $x_i = \{x_i^h\}_{h=1}^s $ and $x_i^h = \{x_{ij}^h\}_{j=1}^{d_h}$, 
the number of clusters $c$, 
the number of views $s$, 
$\alpha$, 
$t=0$, and 
$\varepsilon > 0$.}
\Output{$\mu_{ik}$, $a_{kj}^h$, $v_h$ and clustering labels for input multi-view data $X$.}
$A^{h^{(0)}} = [a_{kj}^{h^{(0)}}]_{c \times d_h}$ and $V^{(0)} = [v_h^{(0)}]_{1 \times s}$ with $v_h = \frac{1}{s} ~\forall h$\;
\For{$\text{iteration} = 1,...,n_{\text{init}}$}{
$\textbf{A} \leftarrow$ InitialCentroids()\;
$\textbf{V} \leftarrow$ InitialWeightedview()\;
$\textbf{A}_{\text{old}} \leftarrow$ $A^{h^{(t-1)}}$\;
$\textbf{V}_{\text{old}} \leftarrow$ $V^{(t-1)}$\;
\For{$t =$ 1,..., $\text{max}_{\text{global}}$}{
    Compute the kernel coefficients $\delta_{ij}^h$\;	
    Compute the memberships $U^{(t)}$ using $A^{h^{(t-1)}}$ and $V_h^{(t-1)}$ by Eq. \ref{eqn:UpdateU_E-MVFMC}\;
    Update the cluster centers $A^{h^{(t)}}$ using $U^{(t)}$ and $V_h^{(t-1)}$ by Eq. \ref{eqn:UpdateA_E-MVFCM}\;
    Update the view weights $V_h^{(t)}$ using $U^{(t)}$ and $A^{h^{(t)}}$ by Eq. \ref{eqn:UpdateV_E-MVFCM}\;
    $\text{movement} \|J_{\text{E-MVFCM}}^{h^{(t)}} - J_{\text{E-MVFCM}}^{h^{(t-1)}}\| < \varepsilon$ \;
    $\textbf{A}, \textbf{A}_{\text{old}}  \leftarrow \textbf{A}_{\text{new}}, \textbf{A}$\;
    $\textbf{V}, \textbf{V}_{\text{old}}  \leftarrow \textbf{V}_{\text{new}}, \textbf{V}$\;
    \uIf{$t \geq \text{max}_{\text{global}}$ or $\text{movement} < \varepsilon$}{
     score $\leftarrow$ RequestScore($(X_{i})_{i \in [n]}$, $\textbf{A} $)\;
     \uIf{score $<$ best\_score}{
        best\_score, best\_centroids $\leftarrow$ score, $\textbf{A}$\;
     }
    \textbf{break}\;
    }
}
}
\textbf{return} best\_centroids\;
\end{algorithm}

\subsection{The EB-MVFCM}

\begin{equation}
J_{\text{EB-MVFCM}} (V, W, U, A) = \sum_{h=1}^s v_h^{\alpha} \sum_{i=1}^n \sum_{k=1}^c (\mu_{ik}^*)^m \|w^h\|^\beta d_{\text{exp}(ik,j)}^h 
\label{eqn:EB-MVFMC}
\end{equation}

\begin{equation*}
\text{s.t.} \sum_{k=1}^c \mu_{ik}^* = 1, \mu_{ik}^* \in [0,1], \sum_{j=1}^{d_h} w_j^h = 1, w_j^h \in [0,1], \text{~~and~~} \sum_{h=1}^s v_h = 1, v_h \in [0,1]
\end{equation*}

Unlike E-MVFCM,  EB-MVFCM objective function integrates the weighted features in the $h$-th view $w_j^h$. $w_j^h$ is used to measure the importances of feature component for each view and regularized by an exponent parameter $\beta$. As visible in Eq. \ref{eqn:EB-MVFMC}, the optimization of EB-MVFCM will involve iteratively updating the "common" memberships $U$, the weight factors $V$, the feature weights $W$, and the cluster centers $A$. Like E-MVFCM, the proposed EB-MVFCM in Eq. \ref{eqn:EB-MVFMC} assigning each data sample to a cluster based on the combined information from multiple views. This is because we believe by enforcing consensus memberships among views in exponential distance strategy can achieve meaningful clustering solutions across data views.

\subsubsection{Optimization}

The Lagrangian of proposed EB-MVFCM in Eq.  \ref{eqn:EB-MVFMC}  can be expressed as:

\begin{equation}
\tilde{J}_{\text{EB-MVFCM}} (V, W, U, A, \lambda_1, \lambda_2, \lambda_3) = J_{\text{EB-MVFCM}} (V, W, U, A) + \lambda_1 (\sum_{k=1}^c \mu_{ik}^* -1) + \lambda_2 (\sum_{h=1}^s v_h -1) + \lambda_3 (\sum_{j=1}^{d_h} w_j^h -1)
\label{eqn:LagEB-MVFMC}
\end{equation}

\paragraph{Step 1:} The updating function for \textit{"common"} memberships $\mu_{ik}^*$

\begin{align*}
\frac{\partial \tilde{J}_{\text{EB-MVFCM}} (V, U, A, \lambda_1, \lambda_2,\lambda_3) }{\partial \mu_{ik}^*} & =  m(\mu_{ik}^*)^{m-1} \sum_{h=1}^s v_h^\alpha \|w^h\|^\beta  d_{\text{exp}(ik,j)}^h  + \lambda_1 = 0\\
m(\mu_{ik}^*)^{m-1} \sum_{h=1}^s v_h^\alpha \|w^h\|^\beta  d_{\text{exp}(ik,j)}^h &=  -\lambda_1 \\
\mu_{ik}^*&= \big(\frac{-\lambda_1}{m}\big)^{\frac{1}{(m-1)}} \bigg( \frac{1}{  \sum_{h=1}^s v_h^\alpha \|w^h\|^\beta   d_{\text{exp}(ik,j)}^h} \bigg)^{\frac{1}{(m-1)}}
\end{align*}

Since $\sum_{k=1}^c \mu_{ik}^* = 1$, then we have $\sum_{k'=1}^c \big(\frac{-\lambda_1}{m}\big)^{\frac{1}{(m-1)}} \bigg( \frac{1}{  \sum_{h=1}^s v_h^\alpha \|w^h\|^\beta d_{\text{exp}(ik',j)}^h} \bigg)^{\frac{1}{(m-1)}} = 1$ and $ \big(\frac{-\lambda_1}{m}\big)^{\frac{1}{(m-1)}} = 1/ \sum_{k'=1}^c \bigg( \frac{1}{\sum_{h=1}^s v_h^\alpha \|w^h\|^\beta d_{\text{exp}(ik',j)}^h} \bigg)^{\frac{1}{(m-1)}} $. Thus we can update the cluster memberships by using the following equation.

\begin{equation}
\mu_{ik}^* = \frac{ \bigg( \frac{1}{\sum_{h=1}^s v_h^\alpha \|w^h\|^\beta d_{\text{exp}(ik,j)}^h} \bigg)^{\frac{1}{(m-1)}}} {\sum_{k'=1}^c \bigg( \frac{1}{\sum_{h=1}^s v_h^\alpha \|w^h\|^\beta d_{\text{exp}(ik',j)}^h} \bigg)^{\frac{1}{(m-1)}}}
\label{eqn:UpdateU_EB-MVFMC}
\end{equation}

\paragraph{Step 2:} The updating function for weight factors  $v_h$
\begin{align*}
\frac{\partial \tilde{J}_{\text{EB-MVFCM}} (V, W, U, A, \lambda_1, \lambda_2, \lambda_3) }{\partial v_h} & =  \alpha v_h^{\alpha - 1} \sum_{i=1}^n \sum_{k=1}^c (\mu_{ik}^*)^m \|w^h\|^\beta d_{\text{exp}(ik,j)}^h + \lambda_2 = 0\\
 v_h^{\alpha - 1} \sum_{i=1}^n \sum_{k=1}^c (\mu_{ik}^*)^m \|w^h\|^\beta d_{\text{exp}(ik,j)}^h &=  -\lambda_2 \\
v_h&= \big(\frac{-\lambda_2}{\alpha}\big)^{\frac{1}{(\alpha - 1)}} \bigg( \frac{1}{   \sum_{i=1}^n \sum_{k=1}^c (\mu_{ik}^*)^m \|w^h\|^\beta d_{\text{exp}(ik,j)}^h } \bigg)^{\frac{1}{(\alpha - 1)}}
\end{align*}

Since $\sum_{h=1}^s v_h = 1$, then we have $\sum_{h'=1}^s \big(\frac{-\lambda_2}{\alpha}\big)^{\frac{1}{(\alpha - 1)}} \bigg( \frac{1}{   \sum_{i=1}^n \sum_{k=1}^c (\mu_{ik}^*)^m \|w^{h'}\|^\beta d_{\text{exp}(ik,j)}^{h'} } \bigg)^{\frac{1}{(\alpha - 1)}} =1$ and $\big(\frac{-\lambda_2}{\alpha}\big)^{\frac{1}{(\alpha - 1)}} = \frac{1} {\sum_{h'=1}^s \bigg( \frac{1}{   \sum_{i=1}^n \sum_{k=1}^c (\mu_{ik}^*)^m \|w^{h'}\|^\beta d_{\text{exp}(ik,j)}^{h'} } \bigg)^{\frac{1}{(\alpha - 1)}} }$. Thus, we can obtain an iterative procedure for updating \textit{h}-th view cluster centers as follows.

\begin{equation}
v_h = \frac{\bigg( \frac{1}{   \sum_{i=1}^n \sum_{k=1}^c (\mu_{ik}^*)^m \|w^h\|^\beta d_{\text{exp}(ik,j)}^h } \bigg)^{\frac{1}{(\alpha - 1)}}} {\sum_{h'=1}^s \bigg( \frac{1}{   \sum_{i=1}^n \sum_{k=1}^c (\mu_{ik}^*)^m \|w^{h'}\|^\beta d_{\text{exp}(ik,j)}^{h'} } \bigg)^{\frac{1}{(\alpha - 1)}}}
\label{eqn:UpdateV_EB-MVFCM}
\end{equation}

\paragraph{Step 3:} The updating function for feature factors in the \textit{h}-th view $w_j^h$
\begin{align*}
\frac{\partial \tilde{J}_{\text{EB-MVFCM}} (V, W, U, A, \lambda_1, \lambda_2, \lambda_3) }{\partial w_j^h} & = \beta (w_j^h)^{\beta -1} v_h^{\alpha} \sum_{i=1}^n \sum_{k=1}^c (\mu_{ik}^*)^m d_{\text{exp}(ik,j)}^h + \lambda_3 = 0\\
\beta (w_j^h)^{\beta -1} v_h^{\alpha} \sum_{i=1}^n \sum_{k=1}^c (\mu_{ik}^*)^m d_{\text{exp}(ik,j)}^h &=  -\lambda_3 \\
w_j^h&= \big(\frac{-\lambda_3}{\beta}\big)^{\frac{1}{(\beta- 1)}} \bigg( \frac{1}{  v_h^{\alpha} \sum_{i=1}^n \sum_{k=1}^c (\mu_{ik}^*)^m d_{\text{exp}(ik,j)}^h  } \bigg)^{\frac{1}{(\beta - 1)}}
\end{align*}

Since $\sum_{j=1}^{d_h} w_j^h = 1$, then we have $\sum_{j'=1}^{d_h} \big(\frac{-\lambda_3}{\beta}\big)^{\frac{1}{(\beta- 1)}} \bigg( \frac{1}{  v_h^{\alpha} \sum_{i=1}^n \sum_{k=1}^c (\mu_{ik}^*)^m d_{\text{exp}(ik,j')}^h  } \bigg)^{\frac{1}{(\beta - 1)}}
 =1$ and $ \big(\frac{-\lambda_3}{\beta}\big)^{\frac{1}{(\beta- 1)}}  = \frac{1} {\sum_{j'=1}^{d_h} \bigg( \frac{1}{  v_h^{\alpha} \sum_{i=1}^n \sum_{k=1}^c (\mu_{ik}^*)^m d_{\text{exp}(ik,j')}^h  } \bigg)^{\frac{1}{(\beta - 1)}} }$. Thus, we can obtain an iterative procedure for updating \textit{h}-th view feature factors $w_j^h$ as follows.

\begin{equation}
w_j^h = \frac{\bigg( \frac{1}{  v_h^{\alpha} \sum_{i=1}^n \sum_{k=1}^c (\mu_{ik}^*)^m d_{\text{exp}(ik,j)}^h  } \bigg)^{\frac{1}{(\beta - 1)}}} {\sum_{j'=1}^{d_h} \bigg( \frac{1}{  v_h^{\alpha} \sum_{i=1}^n \sum_{k=1}^c (\mu_{ik}^*)^m d_{\text{exp}(ik,j')}^h  } \bigg)^{\frac{1}{(\beta - 1)}}}
\label{eqn:UpdateW_EB-MVFCM}
\end{equation}

\paragraph{Step 4:} The updating function for the $h$-th view cluster centers $a_{kj}^h$

\begin{align*}
\frac{\partial J_{\text{EB-MVFCM}} (V, W, U, A) }{\partial a_{kj}^h} & = -2 (x_{ij}^h - a_{kj}^h) v_h^\alpha \sum_{i=1}^n  (\mu_{ik}^*)^m \|w^h\|^\beta \{ - \text{exp}(-\delta_{i}^h \|x_i^h - a_k^h\|^2 \} = 0\\
 v_h^\alpha \sum_{i=1}^n  (\mu_{ik}^*)^m \|w^h\|^\beta \{ - \text{exp}(-\delta_{i}^h \|x_i^h - a_k^h\|^2 \}  a_{kj}^h&=  v_h^\alpha \sum_{i=1}^n  (\mu_{ik}^*)^m \|w^h\|^\beta \{ - \text{exp}(-\delta_{i}^h \|x_i^h - a_k^h\|^2 \}  x_{ij}^h \\
 a_{kj}^h&=  \frac{v_h^\alpha \sum_{i=1}^n  (\mu_{ik}^*)^m \|w^h\|^\beta \{ - \text{exp}(-\delta_{i}^h \|x_i^h - a_k^h\|^2 \} } {v_h^\alpha \sum_{i=1}^n  (\mu_{ik}^*)^m \|w^h\|^\beta \{ - \text{exp}(-\delta_{i}^h \|x_i^h - a_k^h\|^2 \} }  x_{ij}^h 
\end{align*}

Thus, we can update the cluster centers using the following equation.
\begin{equation}
 a_{kj}^h =  \frac{ \sum_{i=1}^n v_h^\alpha (\mu_{ik}^*)^m \|w^h\|^\beta \{ - \text{exp}(-\delta_{i}^h \|x_i^h - a_k^h\|^2 \} } {\sum_{i=1}^n v_h^\alpha (\mu_{ik}^*)^m \|w^h\|^\beta \{ - \text{exp}(-\delta_{i}^h \|x_i^h - a_k^h\|^2 \} }  x_{ij}^h 
 \label{eqn:UpdateA_EB-MVFCM}
\end{equation}

The algorithm for proposed \texttt{EB-MVFCM} in Eq. \ref{eqn:EB-MVFMC} is provided as below.

\begin{algorithm}[h]
\caption{The EB-MVFCM}
\label{alg:EB-MVFCM}
\SetKwInOut{Input}{Input}\SetKwInOut{Output}{Output}
\SetKwProg{ForAllInParallel}{foreach}{ do in parallel}{end} 
\Input{
Multi-view data set $X = \{x_1, x_2, \ldots, x_n\}$ with $x_i = \{x_i^h\}_{h=1}^s $ and $x_i^h = \{x_{ij}^h\}_{j=1}^{d_h}$, 
the number of clusters $c$, 
the number of views $s$, 
$\alpha$, 
$t=0$, and 
$\varepsilon > 0$.}
\Output{$\mu_{ik}$, $a_{kj}^h$, $v_h$ and clustering labels for input multi-view data $X$.}
$A^{h^{(0)}} = [a_{kj}^{h^{(0)}}]_{c \times d_h}$,  $W{h^{(0)}} = [w_{j}^{h^{(0)}}]_{1 \times d_h}$ with $w_j^h = \frac{1}{d_h} ~\forall j$, and $V^{(0)} = [v_h^{(0)}]_{1 \times s}$ with $v_h = \frac{1}{s} ~\forall h$\;
\For{$\text{iteration} = 1,...,n_{\text{init}}$}{
$\textbf{A} \leftarrow$ InitialCentroids()\;
$\textbf{W} \leftarrow$ InitialFeatureweights()\;
$\textbf{V} \leftarrow$ InitialWeightedview()\;
$\textbf{A}_{\text{old}} \leftarrow$ $A^{h^{(t-1)}}$\;
$\textbf{W}_{\text{old}} \leftarrow$ $W^{h^{(t-1)}}$\;
$\textbf{V}_{\text{old}} \leftarrow$ $V^{(t-1)}$\;
\For{$t =$ 1,..., $\text{max}_{\text{global}}$}{
    Compute the kernel coefficients $\delta_{ij}^h$\;	
    Compute the memberships $U^{(t)}$ using $A^{h^{(t-1)}}$ and $V_h^{(t-1)}$ by Eq. \ref{eqn:UpdateU_EB-MVFMC}\;
    Update the cluster centers $A^{h^{(t)}}$ using $U^{(t)}$ and $V_h^{(t-1)}$ by Eq. \ref{eqn:UpdateA_EB-MVFCM}\;
    Update the view weights $V_h^{(t)}$ using $U^{(t)}$ and $A^{h^{(t)}}$ by Eq. \ref{eqn:UpdateV_EB-MVFCM}\;
    Update the feature weights $W^{h^{(t)}}$ using $U^{(t)}$, $V_h^{(t)}$, and $A^{h^{(t)}}$ by Eq. \ref{eqn:UpdateW_EB-MVFCM}\;
    $\text{movement} \|J_{\text{EB-MVFCM}}^{h^{(t)}} - J_{\text{EB-MVFCM}}^{h^{(t-1)}}\| < \varepsilon$ \;
    $\textbf{A}, \textbf{A}_{\text{old}}  \leftarrow \textbf{A}_{\text{new}}, \textbf{A}$\;
    $\textbf{V}, \textbf{V}_{\text{old}}  \leftarrow \textbf{V}_{\text{new}}, \textbf{V}$\;
    $\textbf{W}, \textbf{W}_{\text{old}}  \leftarrow \textbf{W}_{\text{new}}, \textbf{W}$\;
    \uIf{$t \geq \text{max}_{\text{global}}$ or $\text{movement} < \varepsilon$}{
     score $\leftarrow$ RequestScore($(X_{i})_{i \in [n]}$, $\textbf{A} $)\;
     \uIf{score $<$ best\_score}{
        best\_score, best\_centroids $\leftarrow$ score, $\textbf{A}$\;
     }
    \textbf{break}\;
    }
}
}
\textbf{return} best\_centroids\;
\end{algorithm}

\subsection{The significance of the parameter in E-MVFCM and EB-MVFCM}

The parameter settings in proposed E-MVFCM and EB-MVFCM are crucial for optimizing the model's performance and ensuring effective clustering on challenging MV data. Thus, we provide a detailed explanation of the significance of these parameters:

\begin{enumerate}
\item \textbf{The fuzzifier} $m$. This parameter controls the distribution of cluster memberships $\mu_{ik}^*$ in the objective function. In other words, the fuzzifier $m$ helps in exploring the degree of $i$-th sample memberships in all clusters. Proper tuning of $m$ ensures that the model captures the essential structure of MV data without overfitting.
\item \textbf{Exponent parameter} $\alpha$:  This parameter is used to adaptively weight the importance of each view in the clustering process. It helps in balancing the contribution of different views based on their reliability and the amount of available information. Proper tuning of $\alpha$ ensures that the model leverages the most reliable views effectively.
\item \textbf{Exponent parameter} $\beta$: This parameter controls the $j$-th feature weights in the $h$-th view $w_j^h$. Proper tuning of $\beta$ ensures that the model effectively captures the contribution of each feature component in the $h$-th view during the clustering process. 
\item \textbf{The kernel coefficient} $\delta_{ij}^h$: This parameter is crucial to define a kernel-linear with transformation matrices on multi-view environments. For this reason, our proposed methods does not requiring a construction in input data. To set the appropriate $\delta_{ij}^h$, we specify this coefficient by computing the features in the $h$-th view $x_{ij}^h$ and its means $\hat{x}^h$, where $h = 1, \ldots, s$ for $s$ types of heterogeneous features. Here, $\hat{x}^h$ is a constant of the dimension of mass introduced to keep proper dimension of the effective action. Locality of  $\hat{x}^h$ in the objective function helps the transformation of Euclidean distance in kernel space by suppressing non-local contributions. Here, we define local as within one data view. While Non-local as across data views. The aim of this locality of the heat kernel coefficients in the Euclidean distance is to present invariants information across all data views, locally. 
\end{enumerate}

\bibliographystyle{splncs04}  
\bibliography{EB-MVFCM}  

\end{document}